**Contact Tooling Manipulation Control for Robotic Repair Platform – 24508**

Joong-Ku Lee, Young Soo Park
Argonne National Laboratory

## ABSTRACT

This paper delves into various robotic manipulation control methods designed for dynamic contact tooling operations on a robotic repair platform. The explored control strategies include hybrid position-force control, admittance control, bilateral telerobotic control, virtual fixture, and shared control. Each approach is elucidated and assessed in terms of its applicability and effectiveness for handling contact tooling tasks in real-world repair scenarios. The hybrid position-force controller is highlighted for its proficiency in executing precise force-required tasks, but it demands contingent on an accurate model of the environment and structured, static environment. In contrast, for unstructured environments, bilateral teleoperation control is investigated, revealing that the compliance with the remote robot controller is crucial for stable contact, albeit at the expense of reduced motion tracking performance. Moreover, advanced controllers for tooling manipulation tasks, such as virtual fixture and shared control approaches, are investigated for their potential applications.

## INTRODUCTION

DOE-EM has a pressing need for on-site repair and maintenance of the legacy waste management facilities, and deployment of robotic systems is considered to replace human workers from the hazardous tasks. Recently the Washington River Protection System (WRPS) has initiated the use of a robotic platform for valve pit liner repair tasks at the Hanford Tank Farm. The success of this deployment has sparked interest in expanding the utilization of robotic systems for a broader range of repair tasks (Fig. 1). However, the repair and maintenance operations, especially those involving complex contact manipulation of tools, possess significant technical challenges. Recently the emergence of collaborative robots (co-robots) has provided a unique opportunity to address such challenges. Collaborative robots, compared to the traditional industrial robots, possess inherent joint torque sensing capability that can be leveraged to implement dynamic control approaches effective for various contact tooling operations. Argonne National Laboratory (ANL) and Florida International University (FIU) are collaborating on the development of a robotic platform capable of various tooling operations.

This paper entails the development of various robotic manipulation control methods specifically designed for dynamic contact tooling operations. While most non-contact type tooling operations are feasible with conventional industrial robotic controllers, contact tooling operations, such as cutting and grinding, require additional development of dynamic control methods. In this regard, the project explores a comprehensive set of robotic operation modes, including automation, teleoperation, and shared autonomy, to cover a wide range of repair scenarios. The project is still in progress and includes ongoing technology integration and field demonstration. Upon successful integration, the multi-purpose robotic tooling platform is expected to provide a crucial enabling capability for the EM missions.

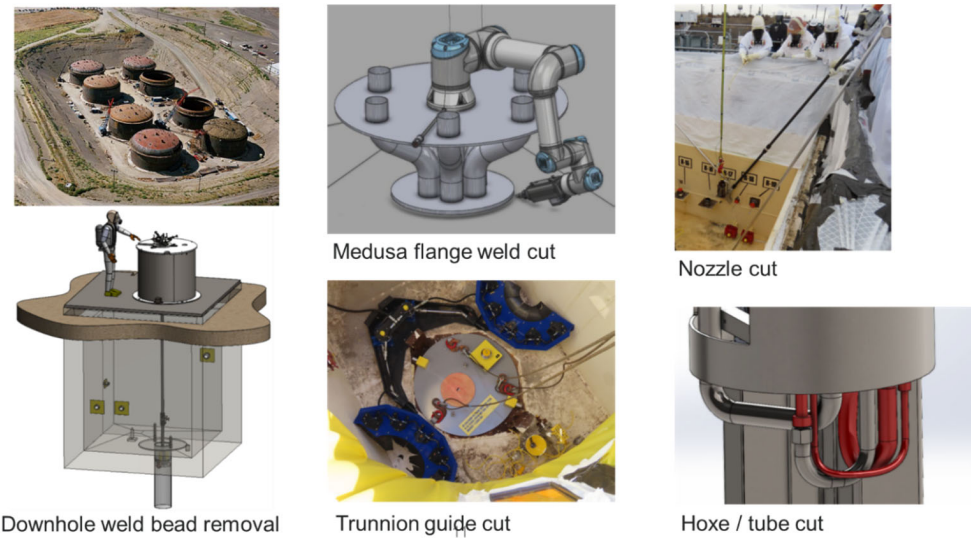

**Figure 1.** Various tank farm repair and maintenance tasks requiring contact tooling that might be implemented by advanced robotics systems.

## SYSTEM OVERVIEW

The robotic system discussed in this paper, equipped with interchangeable tools, is currently in development at Florida International University. This system comprises a collaborative robot arm, specifically the UR16e by Universal Robot, paired with a custom-made wheeled mobile robot base. The robot arm's end-effector is fitted with a plug-and-play tool changer, enabling the adoption of various repair tools, including cutters, grinders, grippers, drills, sprayers, welders, and more. Fig. 2 illustrates the configuration of the repair tooling robot system, which is also controlled via teleoperation.

Anticipated scenarios involve significant contact interactions with the external environment, particularly with tools like cutters and grinders. Additionally, unlike fixed automation in a factory environment, manipulation on a mobile base introduces the challenge of a non-rigid foundation. These factors may contribute to dynamic instability during contact manipulation. To ensure stable contact tooling under such conditions, dynamic manipulation control methods need to be implemented.

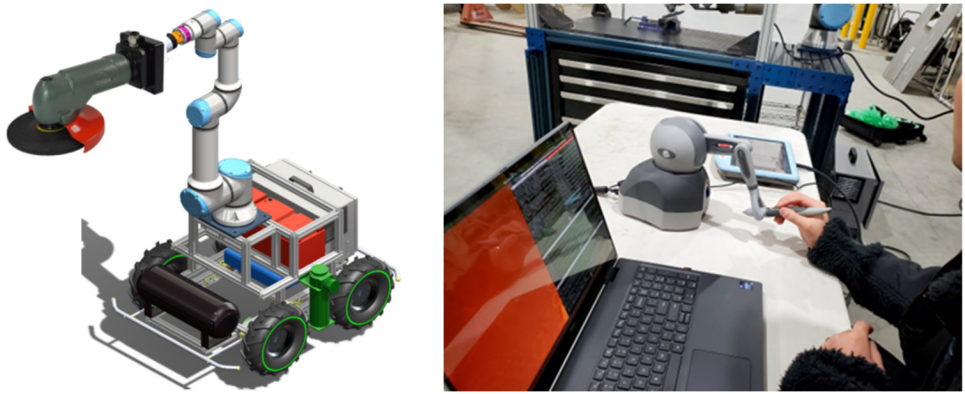

**Figure 2.** Robotic system with plug-and-play interchangeable tools is being developed by Florida International University

**METHODS**

For repair tasks, a crucial requirement is the ability to manipulate contact tools, such as cutting and grinding tools, which involve applying high forces onto work objects while maintaining stable contact. In contact manipulation, effective control of contact dynamics (via impedance control) and force control is necessary to achieve simultaneous stability and precision in force application.

To facilitate effective and stable contact tooling manipulation with the collaborative robotic platform, we have explored the implementation of various dynamic control and manipulation methods, including:

- Hybrid position-force control
- Bilateral telerobotic control
- Virtual fixtures
- Teleautonomy

This section presents the implementation of each method.

**Hybrid Position-Force Control**

In situations where the external environment exhibits structured and static nature, precise control inputs to the robot controller can be computed using external sensors such as computer vision systems. In such cases, hybrid force/position control [1] is a suitable approach for conducting contact manipulation tasks. This approach involves either the user or an automated agent planning the trajectory for manipulation task and determining both the direction and magnitude of the desired force.

Fig. 3 illustrates the block diagram of this control system. Assuming $S$ as a matrix representing the direction of the desired force, the direction of the robot movement can be specified as $I - S$. Consequently, the dynamics of the two controllers can be decoupled by the orthogonal projections $S$ and $I - S$. The resulting expression for the hybrid position-force controller is as follows:

$$V = (I - S) \cdot V_d + S \cdot \left( K_{fp} \cdot F_e + K_{fi} \int F_e(t) dt \right),$$

where $V_d$ is the calculated velocity for the given desired trajectory, $K_{fp}$ is the proportional gain for the force control loop, and $K_{fi}$ is the integral gain for the force control loop. Within the control loop, the robot is force-controlled in the designated direction while being position-controlled in the other direction to maintain the trajectory.

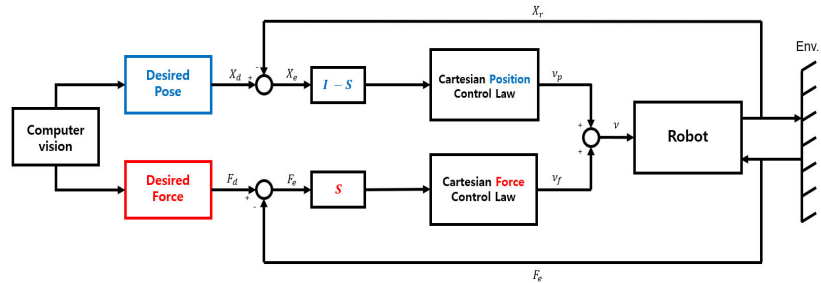

**Figure 3.** The block diagram of hybrid force/position controller

The controller is tested on the UR16e corobot arm for contact manipulation as illustrated in Fig. 4. As shown in Fig. 4the figure, the test result shows that accurate force control was made possible.

While this method demonstrates good performance in contact manipulation tasks, it requires a precise prior knowledge of the tool path and desired force profile during contact manipulation tasks.

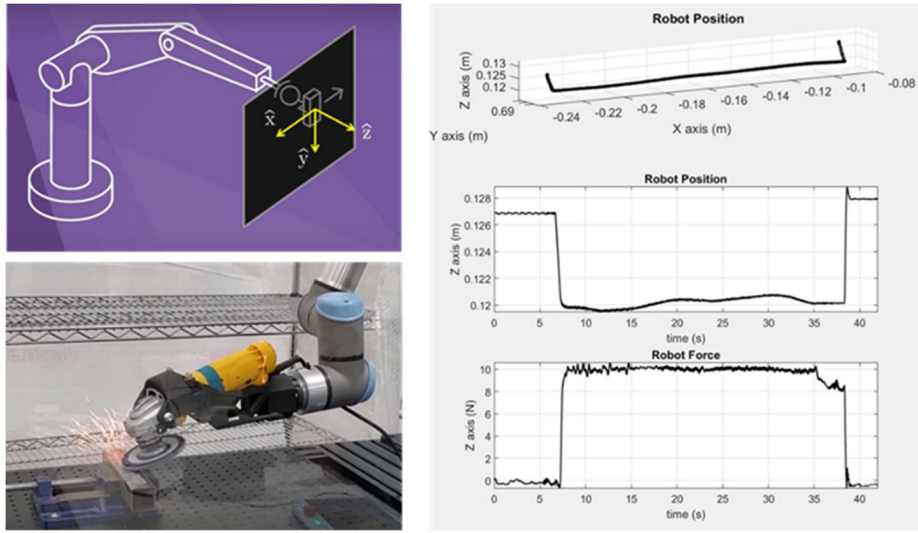

**Figure 4.** Demonstration of the contact tooling task with hybrid force/position controller

**Bilateral Teleoperation Control**

To perform contact manipulation tasks with the robotic system in unstructured or dynamic environments, telerobotic operation may be adopted. In teleoperation, a human operator controls the robot manipulator from a distance using an input device based on sensory feedback. By involving the human operator in the control loop, the system can leverage human perception, planning, and control abilities, even in unpredictable environments.

The ultimate goal of teleoperation is to achieve 'telepresence' via multi-modal perception-action loop, where visual and haptic feedback is the main modality. In particular, bilateral telerobotic control system [2] providing force feedback to human operator is essential. Bilateral control is a complex system which involves dynamic control as well as motion feedback control, as illustrated in Fig. 5. This capability is crucial for maintaining control and adaptability during contact manipulation tasks in such environments.

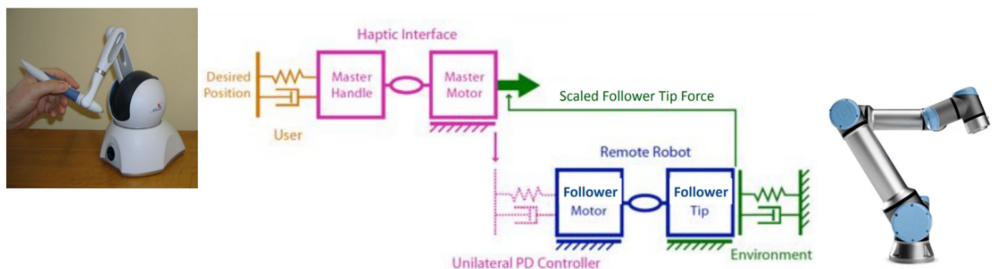

**Figure 5.** Schematic diagram of bilateral telerobotic operation controller. The movement of the haptic interface is transferred to the remote robot, while robot tip force is transferred back to the haptic device.

The bilateral telerobotic control system, encompassing both the robot's interaction dynamics with the environment and perception-action feedback in teleoperation, was assessed in four different configurations:

a)  Position-force bilateral control, Admittance control of robot arm
b)  Position-position bilateral control, Admittance control of robot arm
c)  Position-force bilateral control, position control of robot arm
d)  Position-position bilateral control, position control of robot arm.

Fig. 6 illustrates architectures of both position-position and position-force type bilateral controls. The position controller is implemented to precisely track the given pose from the teleoperation. Therefore, it lacks compliance, which is the ability to yield to external forces or disturbances. In contrast, the admittance controller is implemented to control robot with user-defined compliance.

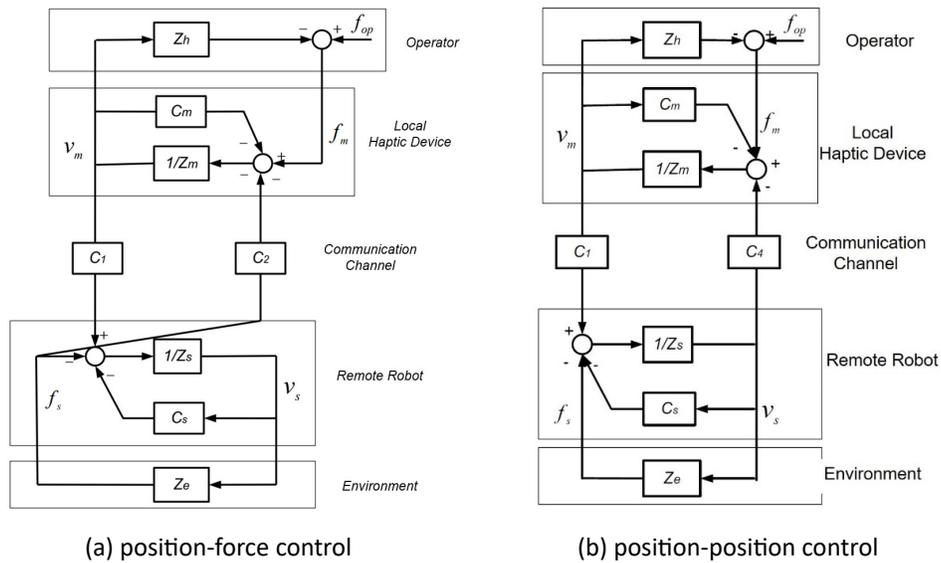

(a) position-force control                    (b) position-position control

**Figure 6.** Bilateral teleoperation control system architectures.

A series of contact manipulation test operations were carried out with various control algorithms and test pieces. As illustrated in Fig. 7, the test operations involved removing weld beads on metal blocks with a metal grinder. The operation was performed in teleoperation.

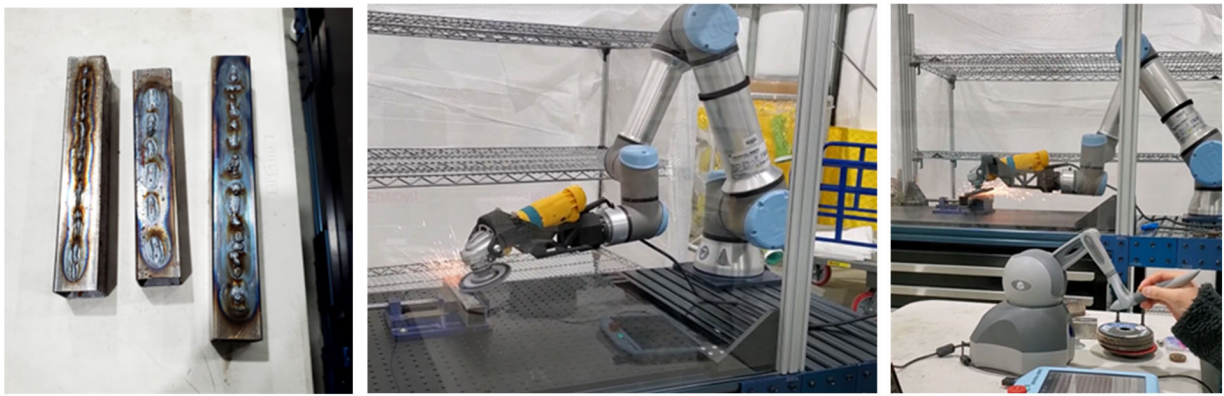

**Figure 7.** Contact manipulation test on flat surface workpiece

The results of the bilateral teleoperation test for weld bead grinding operations on planar workpieces are depicted in Fig. 8. The top row presents the 3-dimensional position of the robot end-effector during the operation, while the bottom row displays the position and force in the z-direction (normal to the plane surface). The force data shown has been smoothed with a 50-point ARMA filter.

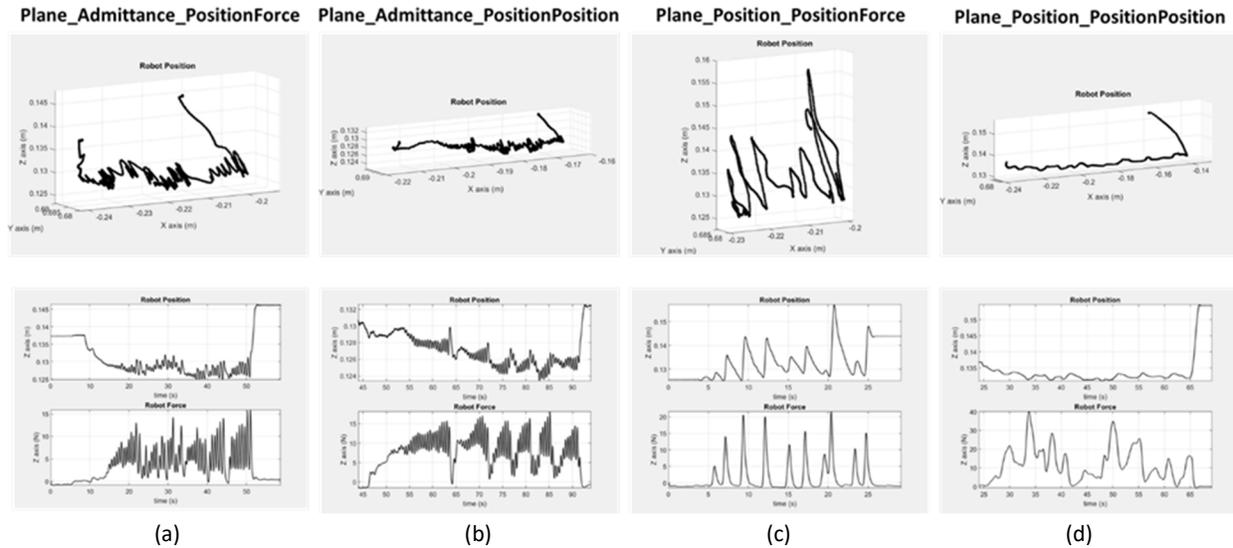

(a)      (b)      (c)      (d)

**Figure 8.** Test Results of Bilateral Teleoperation for Weld Bead Grinding Operation on Planar Work Pieces: Top row shows the 3D tool position, and bottom row shows the z-axis position and force at the robot end-effector.

Analysis of Test Results:

a) In this test employing position-force bilateral control with admittance control of the robot arm, the dynamic responsiveness introduced by admittance control resulted in high-frequency components in the motion data. Despite the noticeable fluctuation, the magnitude was maintained at a manageable level. In this bilateral control setup, the tool contact force was directly reflected to the operator, who felt and utilized it to control the robot. The limited responsiveness of the human hand manifested as the low-frequency (DC) component of the motion data. Although the data appeared noisy, both position and force data indicated the onsets of the weld bead removal processes. Overall, this test demonstrated the effective performance of the weld cut operation with high responsiveness to dynamic interactions.

b) In this test involving position-position bilateral control with admittance control of the robot arm, dynamic responsiveness was introduced by the admittance control, resulting in high-frequency components in the motion data. Similar to the previous test, the fluctuation magnitude was kept at a manageable level. In this type of bilateral control, force reflection was achieved through position feedback, specifically the position difference between the hand-controller and the robot. This setup allowed the human operator to respond adequately. The plots demonstrated that the motion trajectory was well maintained along the desired cutting path, and the contact tooling force was consistently applied during the weld bead removal process. Overall, this test indicated good tooling performance with responsiveness to contact dynamic interactions in the weld cut operation.

c) In this test involving position-position bilateral control with position control of the robot arm, the absence of admittance control led to the disappearance of high-frequency components in the motion data. Without dynamic control, the operator faced challenges in responding adequately. The direct force reflection resulted in inadequate performance in contact manipulation, where any physical contact led to a high build-up of forces and motion, as depicted in the plots. Overall, this test indicated poorer performance for contact manipulation under this setup.

d) In this test involving position-position bilateral control with position control of the robot arm, the absence of admittance control resulted in the disappearance of high-frequency components in the motion data. Despite the lack of dynamic control, under the low-frequency position-type force reflection, the operator was able to respond adequately. The contact manipulation showed good performance, distinctly identifying the onset of physical contacts with beads and maintaining a continuous application of tooling force. Surprisingly, this test result demonstrated the best performance for contact manipulation under the given setup. There was initial suspicion that the simplicity of the workpiece's geometry allowed the operator to navigate without the need to feel the objects and vary motion trajectory commands. Subsequently, it was suggested to perform the test operation with a more complex-shaped workpiece for further evaluation.

**Preliminary Test of Enhanced Teleoperation Methods**

In the previous manual teleoperation test, performance limitations were encountered, primarily stemming from the inherent constraints in human perception-action capabilities. Notably, challenges arose from the restricted ability of humans to perceive environmental geometry and the constrained perception-action bandwidth. To address these limitations effectively, innovative telerobotic operation methods can be developed, leveraging augmented reality technologies and enhanced robot autonomy.

a) *Virtual fixtures*: In this method of teleoperation, we introduce augmented-reality technology, virtual fixtures [5]. to aid human operators in teleoperation. In this approach, artificial geometries, e.g., surfaces, are generated and overlaid onto the human perception, visual and haptic, in such a way as to guide in teleoperation. Providing such virtual fixtures for the operation can feel and use can effectively simplify various teleoperation tasks. Fig. 9 depicts the conceptual diagram of the virtual fixture aided bilateral teleoperation system, where virtual feedback is generated by virtual fixture and applied to the user. Fig. 10 illustrates some example uses of virtual fixtures in various remote operation tasks. In teleoperation, the same types of teleoperation control methods can be adopted, however, while the sensory feedback is from interaction with the virtual fixtures, instead of from the real workpieces. This passive approach will result in always stable operations if the virtual fixture is correctly placed. This method was first invented in the space robotics community, and later improvise for applications in nuclear industry, and surgery robots. This method is known to be effective in tasks in semi-structured, and relative static environments, where the operator can spend much time in designing and placing the virtual fixtures in operator interface. Recently, ongoing reserarches aim to facilitate users in defining or even automating the virtual fixture genereation process for increased user-friendliness [6, 7].

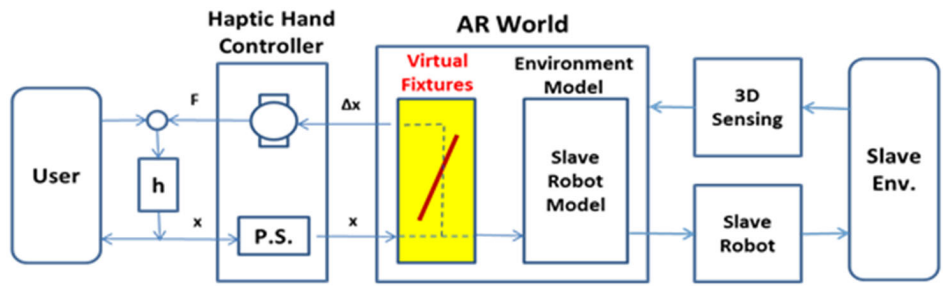

**Figure 9.** Conceptual diagram of the Virtual Fixture aided bilateral teleoperation system.

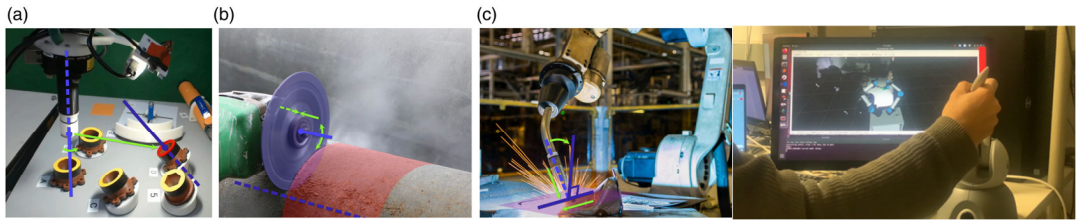

**Figure 10.** Use of virtual fixture for the teleoperation.

Test operation was performed to test the feasibility of utilizing virtual fixtures for the contact tooling method. In this particular test, a planar work piece was employed, and a virtual fixture with planar geometry was applied on top of the workpiece. Notably. The workpiece had minimal weld beads remaining, having been largely consumed in previous test operations. While multi-modal (visual-haptic) virtual fixtures are implemented in general, our test implementation adopted only haptic feedback.

Figure 12(a) presents the test results, demonstrating the capability to maintain a highly precise tool path with minimal fluctuations. Additionally, the tooling force was consistently maintained at a high-quality level. It's crucial to consider that such performance relies on the assumption of a perfectly overlaid virtual fixture.

*b) Shared control (teleautonomy):* Another approach to alleviate the mental and physical workload on human operators is shared control, also known as teleautonomy. This method integrates the task execution skills of the human operator with the learning capabilities of an autonomous agent. Figure 15 provides an illustration of this concept. In this approach, the autonomous agent acquires skills from the human operator's contact manipulation task performance through demonstrations. Subsequently, employing an interactive shared controller, the autonomous agent collaborates with the human operator, applying the learned skills to assist in the execution of contact manipulation tasks. Ongoing research continues to explore and refine shared autonomy, incorporating advancements in AI and machine learning methods.

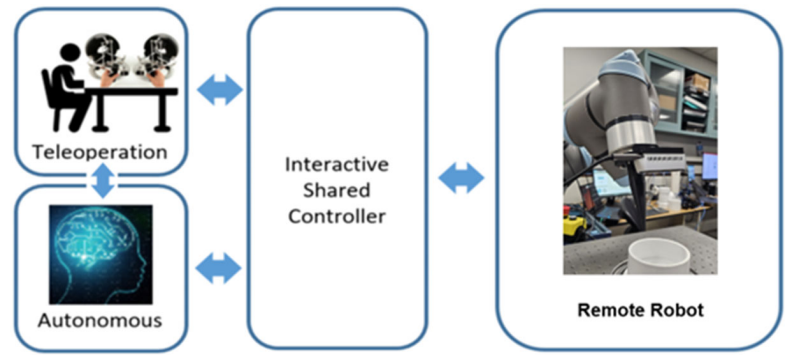

**Figure 11.** Conceptual diagram of shared teleoperation controller

To evaluate the feasibility of implementing shared autonomy for Hanford facility repair operations, we have performed a rudimentary implementation of shared autonomy. In this implementation, we used the planar workpiece, and implemented a teleoperation approach similar to the hybrid position-force control, previously implemented. In this case, while automatic force control is invoked in the normal direction, the lateral motion is left to rely on human operator's manual control.

Fig. 12(b) shows the test results. As can be seen, it was possible to maintain relatively precise planar tool path. Also tooling force was maintained in perfect quality. (However, keep in mind that such performance is based on the assumption that tool path can be identified vis external sensing.)

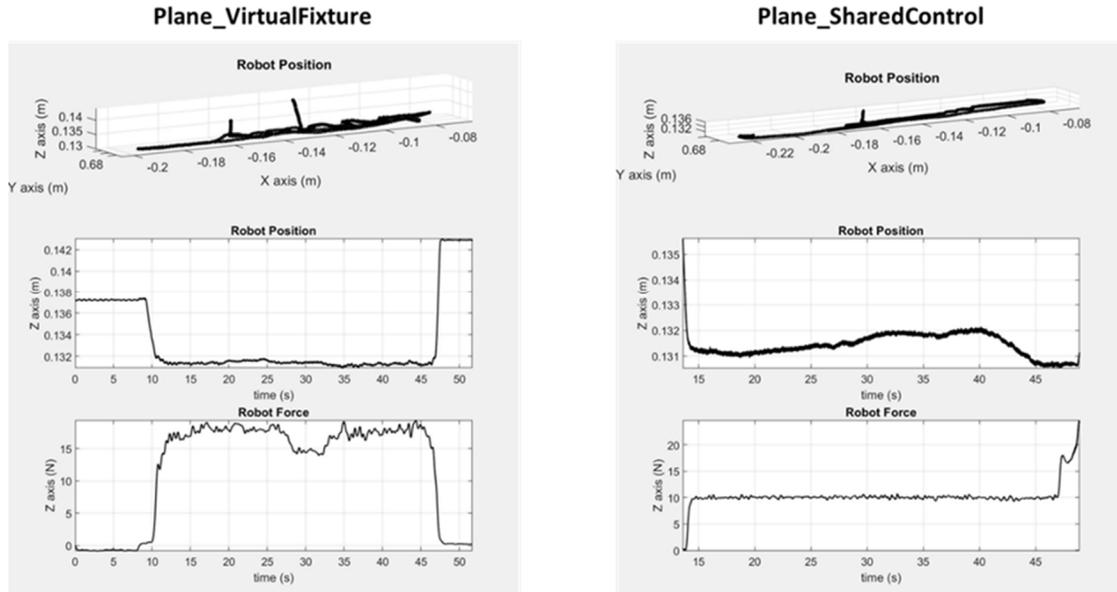

(a) virtual fixtures              (b) shared control

**Figure 12.** Feasibility Test Result of Enhanced Telerobotic Operation

## APPLICATION DEMONSTRATION – DOWNHOLE WELD BEAD REMOVAL

After the technology feasibility test, the WRPS project team introduced a near-real task object (Fig. 13), featuring circumferential welds along the of the inner surface. Utilizing this workpiece, we conducted a test demonstration of downhole weld bead removal operation (Fig. 14). Building on observations from previous test operations, we employed the position-position teleoperation method with an admittance-controlled robot arm. Fig. 15 illustrates the results (3D position and normal force) of the test demonstration, along with an image of the test piece with the weld removed after the operation.

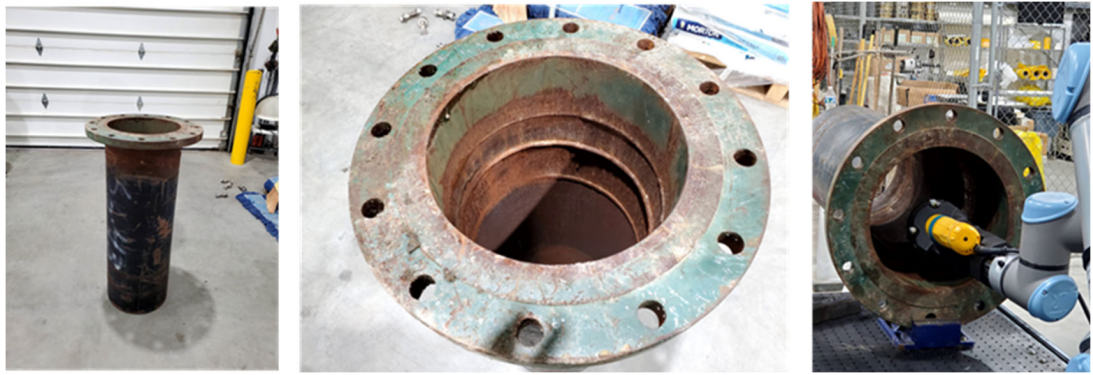

**Figure 13.** Demonstration Test Object with Downhole Weld Beads.

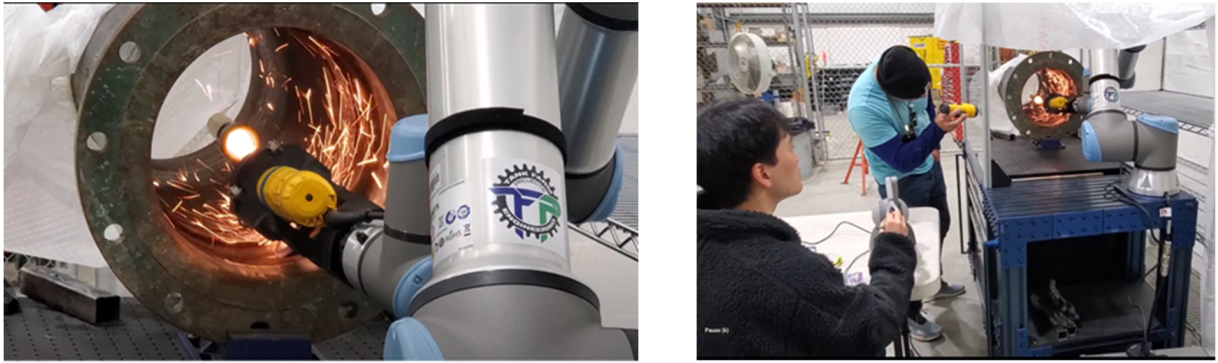

**Figure 14.** Test Teleoperation of Downhole Weld Bead Removal.

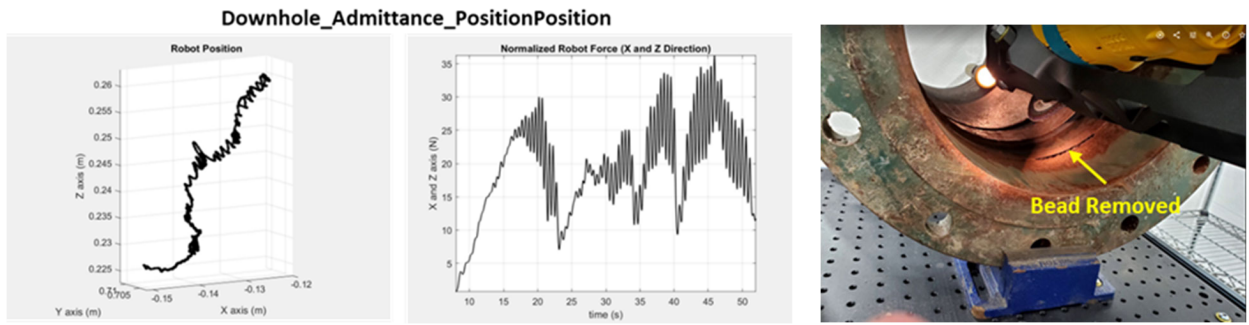

**Figure 15.** Test Demonstration Result of Downhole Weld Bead Removal.

## SUMMARY AND CONCLUSIONS

In this paper, we present the development of various robotic manipulation control methods for dynamic contact tooling operations. While most non-contact type tooling operation are feasible with conventional industrial robotic controllers, many heavy cutting operations require additional development of dynamic control methods. To explore all situations of repair tasks, the development incorporating a comprehensive set of robotic operation modes have been implemented and tested, including hybrid force/position controller, position-based bilateral teleoperation controller, admittance-based bilateral teleoperation controller, virtual fixture, and shared control approaches.

## REFERENCES

1. Lynch, K. M., & Park, F. C. (2017). 11.6 Hybrid Motion-Force Control. In Modern robotics. Cambridge University Press.

2. Kuchenbecker, Katherine Julianne. Characterizing and controlling the high-frequency dynamics of haptic interfaces. Stanford University, 2006.

3. Lawrence, D. A. (1993). Stability and transparency in bilateral teleoperation. IEEE transactions on robotics and automation, 9(5), 624-637.

4. Lynch, K. M., & Park, F. C. (2017). 11.7.2 Admittance-Control Algorithm. In Modern robotics. Cambridge University Press.

5. Abbott, J. J., & Okamura, A. M. (2003, September). Virtual fixture architectures for telemanipulation. In 2003 IEEE International Conference on Robotics and Automation (Cat. No. 03CH37422) (Vol. 2, pp. 2798-2805). IEEE.

6.  Pruks, Vitalii, and Jee-Hwan Ryu. "Method for generating real-time interactive virtual fixture for shared teleoperation in unknown environments." The International Journal of Robotics Research 41.9-10 (2022): 925-951.

7.  Lee, H. J., & Brell-Cokcan, S. (2023). Reinforcement Learning-based Virtual Fixtures for Teleoperation of Hydraulic Construction Machine. arXiv preprint arXiv:2306.11897.

8.  Zeestraten, Martijn JA, Ioannis Havoutis, and Sylvain Calinon. "Programming by demonstration for shared control with an application in teleoperation." IEEE Robotics and Automation Letters 3.3 (2018): 1848-1855.

9.  Michel, Y., Li, Z., & Lee, D. (2023). A Learning-Based Shared Control Approach for Contact Tasks. IEEE Robotics and Automation Letters.

## ACKNOWLEDGEMENTS

This work is supported by the U.S. Department of Energy, Office of Environment Management and Korea Atomic Energy Research Institute. This manuscript has been created by UChicago Argonne, LLC, Operator of Argonne National Laboratory ("Argonne"). Argonne is a U.S. Department of Energy Office of Science laboratory. This work was performed in collaboration with Florida International University and Washington River Protection System.